\documentclass[conference,letterpaper]{IEEEtran}

\usepackage[letterpaper, left=0.75in, right=0.75in, bottom=0.75in, top=0.75in]{geometry}

\usepackage{cite}
\usepackage{amsmath}
\usepackage{amssymb}
\usepackage{amsthm}
\usepackage{array}
\usepackage{graphicx}
\usepackage{caption}
\usepackage{subcaption}
\usepackage{algorithm}
\usepackage{algorithmic}
\usepackage{comment}
\usepackage[utf8]{inputenc}
\usepackage{subcaption}
\usepackage{multirow, tabularx}
\usepackage{booktabs,ragged2e}

\usepackage[flushleft]{threeparttable}

\title{\vspace{0.25in}Vision-Based Shape Reconstruction of Soft Continuum Arms Using a Geometric Strain Parametrization}

\author{\IEEEauthorblockN{Ali AlBeladi\IEEEauthorrefmark{1},
Girish Krishnan\IEEEauthorrefmark{2},
Mohamed-Ali Belabbas\IEEEauthorrefmark{1}, 
Seth Hutchinson\IEEEauthorrefmark{3},}
\IEEEauthorblockA{\IEEEauthorrefmark{1}Electrical and Computer Engineering, \IEEEauthorrefmark{2}Industrial and Enterprise Systems Engineering,\\ University of Illinois at Urbana-Champaign, Urbana, IL
\\ \{albelad2, gkrishna, belabbas\}@illinois.edu \\
\IEEEauthorrefmark{3}School of Interactive Computing, Georgia Institute of Technology, Atlanta, GA
\\seth@gatech.edu}}

\date{April 2020}

\begin{document}

\maketitle
\begin{abstract}
    Interest in soft continuum arms has increased as their inherent material elasticity enables safe and adaptive interactions with the environment. However to achieve full autonomy in these arms, accurate three-dimensional shape sensing is needed. Vision-based solutions have been found to be effective in estimating the shape of soft continuum arms. In this paper, a vision-based shape estimator that utilizes a geometric strain based representation for the soft continuum arm's shape, is proposed. This representation reduces the dimension of the curved shape to a finite set of strain basis functions, thereby allowing for efficient optimization for the shape that best fits the observed image. Experimental results demonstrate the effectiveness of the proposed approach in estimating the end effector with accuracy less than the soft arm's radius. Multiple basis functions are also analyzed and compared for the specific soft continuum arm in use.
\end{abstract}

\section{Introduction}
Bioinspired soft robots \cite{rus2015design} use stretchable skins, muscles, fluids, fibers and tendons to deform in a continuum fashion. Soft Continuum Arms (SCAs) \cite{Cianchetti2013,octarm2007,Uppalapati2021} are long and slender soft robots inspired by octopus arms and elephant trunks, and possess large spatial workspace and dexterity. The material elasticity combined with inherent damping enables safe and adaptable interaction with the surroundings. SCAs can be useful for several applications such as surgery \cite{Cianchetti2013}, agriculture \cite{Uppalapati2020a}, search and rescue \cite{Rahn2} etc.

Though useful, the curvilinear nature of deformation precludes traditional sensing methods such as encoders. Methods that are specific to sense continuum deformation such as Fiber Bragg Grating \cite{Xu2016} and electromagnetic sensors are effective, but interfere with the felxibility of the SCA or are altered by environmental disturbances \cite{Shi17}. Alternatively, vision-based sensing \cite{Han03,Han05,Chi07,Rei11,Rei12,Cam08,Cro10,Man19,Jac16,Gre18,SXu18,Che19,Rei12b,Rei13,Cab14,Cab17} has gained prominence as it is noninvasive and easy to implement. Vision-based sensing methods seek to reconstruct the 3D shape of the SCA from images obtained from a camera placed in close proximity to the arm.

\begin{figure}
    \centering
    \begin{subfigure}[b]{0.48\textwidth}
         \centering
         \includegraphics[width=\textwidth]{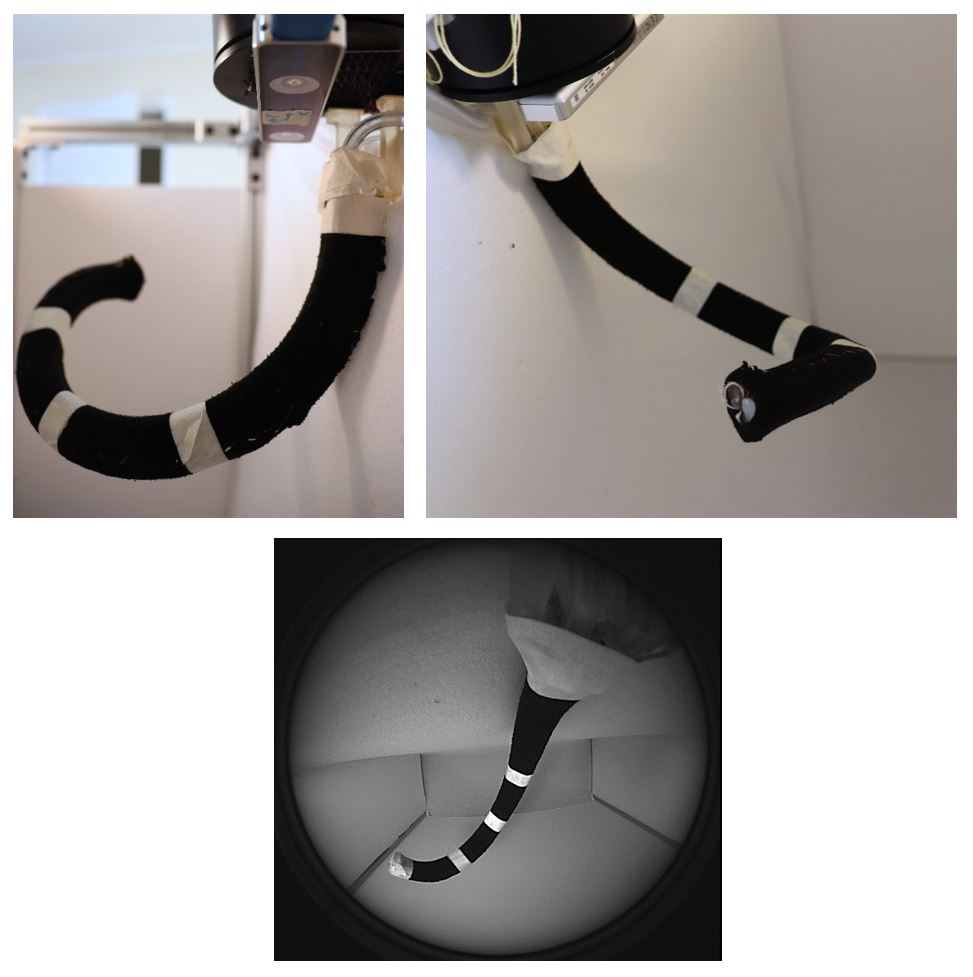}
    \end{subfigure}

    \caption{(above) Images of the $BR^2$ soft continuum arm \cite{Uppalapati2021} from different angles showing a configuration that involves simultaneous bending and twist. (bottom) The view from the camera used for estimating the shape.}
    \label{fig:exp_setup}
\end{figure}



Early efforts on shape reconstruction were limited to planar deformations with the camera placed perpendicular to the SCA's bending plane \cite{Han03,Han05}. A two dimensional shape was estimated by first extracting the visual markers on the SCA from the image and then fitting a two-dimensional curve with piece-wise constant curvatures to these points. 
Recently, Fan et al. \cite{Fan20} proposed to replace the constant curvature with a linearly varying curvature that was fitted through image points of a SCA. The 3D shape of deformable objects was reconstructed in \cite{SXu18} by  applying Self-Organized Maps (SOM) on point cloud data coming from a depth camera. Methods relying on triangulation of multiple cameras were also proposed using the shape-from-silhouette \cite{Cam08}, modified SOM \cite{Cro10}, and learning-based methods \cite{Rei11,Rei12}. Although these methods performed desirably, they require multiple
cameras or depth cameras, which is not always practical, such as in endoscopic surgery \cite{Cab14}.
To solve this, several methods have been proposed to estimate the shape of a flexible endoscopic instrument through optimizing for the minimum image reprojection error \cite{Rei12b,Rei13,Cab14,Cab17}. The model these methods used was specifically for flexible tools in endoscopic surgery and do not apply to other types of continuum manipulators. The work presented in this paper generalizes these methods to any SCA, as long as it is within the field of view of the camera.

In this paper, we propose a geometric strain parameterization method to reconstruct the 3D shape of a fixed-length SCA using a wide-angle monocular camera. 
The shape of the soft arm is represented as a linear combination of a specified set of strain basis functions. Though a similar representation has been introduced in \cite{Ren20} to model soft arms, this paper investigates it in the context of shape reconstruction.
The geometric strain parameterization allows for reconstructing the orientations of the SCA's cross sections along its length as well as the full 3D shape. 
By adopting this representation, the dimensionality of the problem is reduced, thereby easing the estimation process, and making it possible to optimize for the shape that best fits the camera's observations.
We also provide a simple comparison of various basis functions based on their respective accuracy in sensing shape. Although not comprehensive, this comparison provides some insight on what basis performs best for the SCA used in this study. 
    
The proposed shape reconstruction method works as follows. Given initial curvatures along the SCA, the projection of the shape onto the image space is first obtained. Section \ref{sec:background} elaborates the forward model that gives us the image projection of the SCA given its curvature. Then the error between this estimated projection and the observed projection in the camera output is measured. An optimization routine then updates the strain coefficient estimates in a direction that decreases this error. This is repeated until the desired stopping criteria is reached. More details on the approach are presented in Section \ref{sec:shape_recons}.
The method is tested on a fiber reinforced SCA known as the $BR^2$ manipulator \cite{Uppalapati2021} (shown in Fig. \ref{fig:exp_setup}) that can undergo complex bending and twisting configurations. Discussion of the results is presented in Section \ref{sec:results}.

\begin{figure}
    \centering
    \includegraphics[width = \columnwidth]{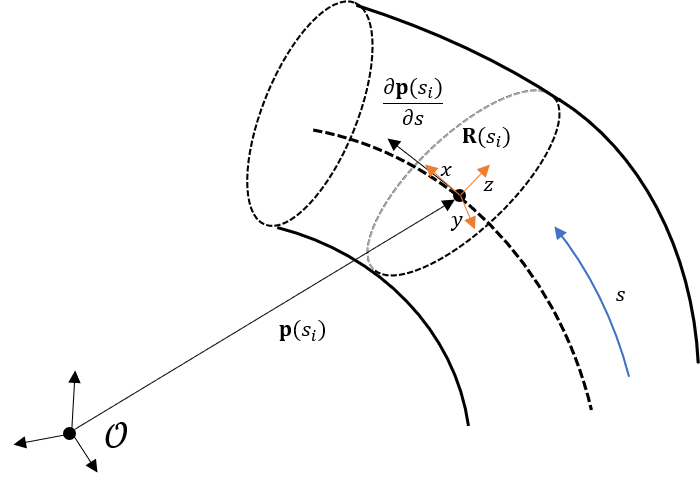}
    \caption{An illustration of the SCA and the corresponding position $\mathbf{p}$, orientation $\mathbf{R}$, and tangent $\frac{\partial{\mathbf{p}}}{\partial{s}}$ of a single cross-section.}
    \label{fig:soft_robot_illustration}
\end{figure}

\begin{figure*}[ht]
     \centering
     \begin{subfigure}[b]{0.19\textwidth}
         \centering
         \includegraphics[width=\textwidth]{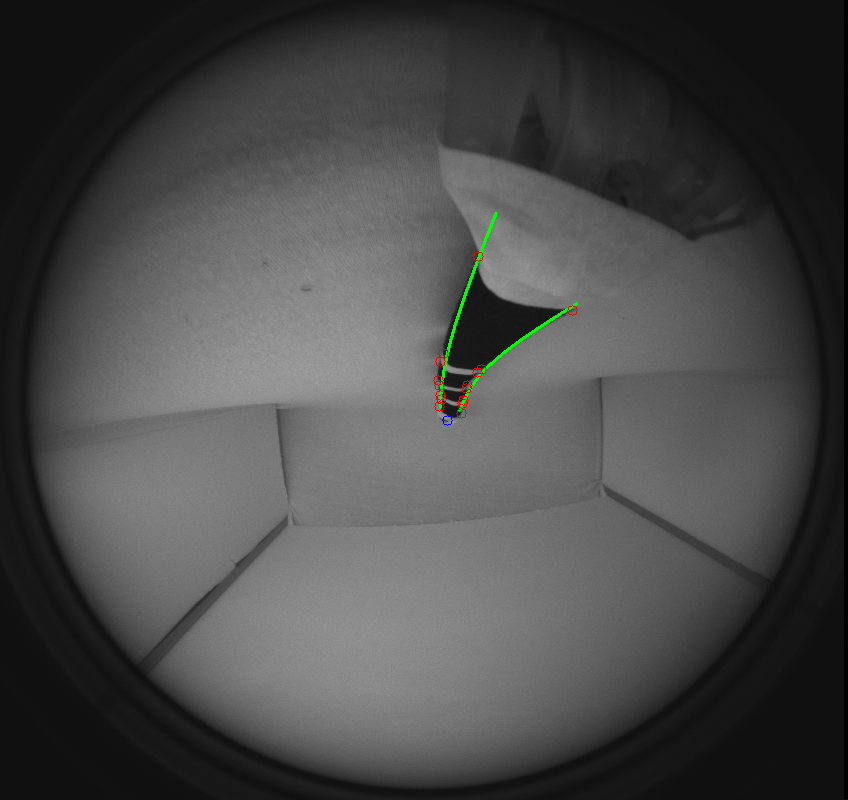}
     \end{subfigure}
     \begin{subfigure}[b]{0.19\textwidth}
         \centering
         \includegraphics[width=\textwidth]{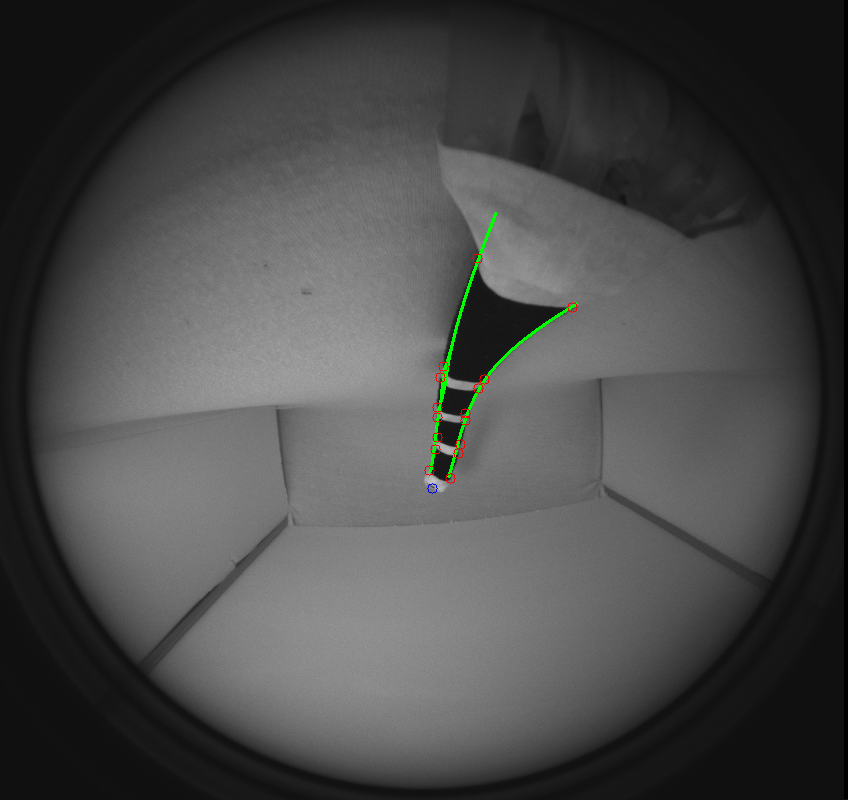}
     \end{subfigure}
     \begin{subfigure}[b]{0.19\textwidth}
         \centering
         \includegraphics[width=\textwidth]{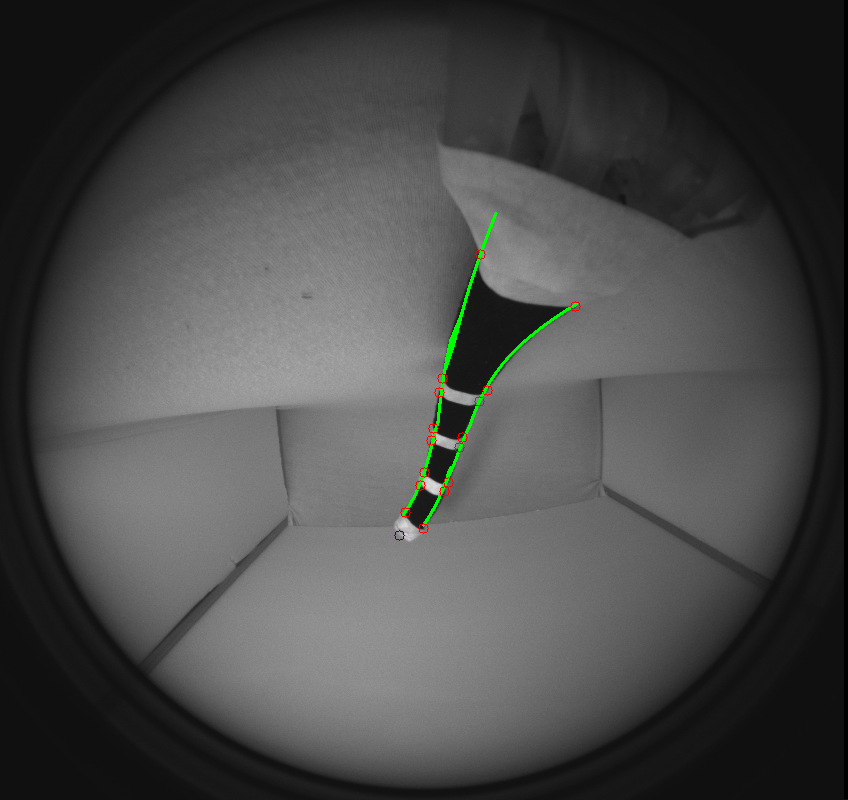}
     \end{subfigure}
     \begin{subfigure}[b]{0.19\textwidth}
         \centering
         \includegraphics[width=\textwidth]{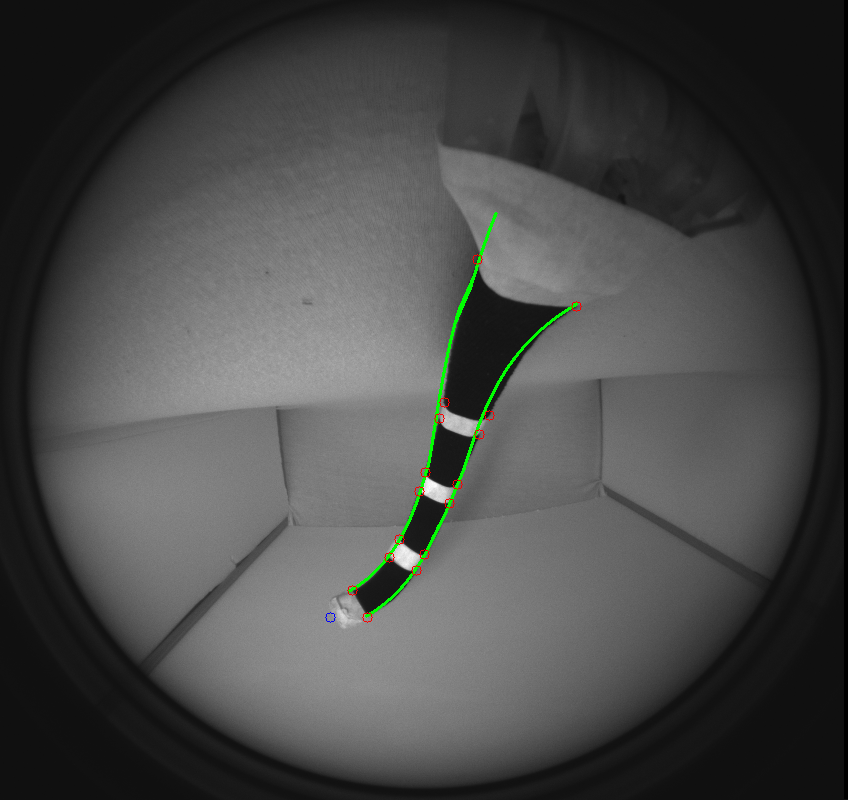}
     \end{subfigure}
      \begin{subfigure}[b]{0.19\textwidth}
         \centering
         \includegraphics[width=\textwidth]{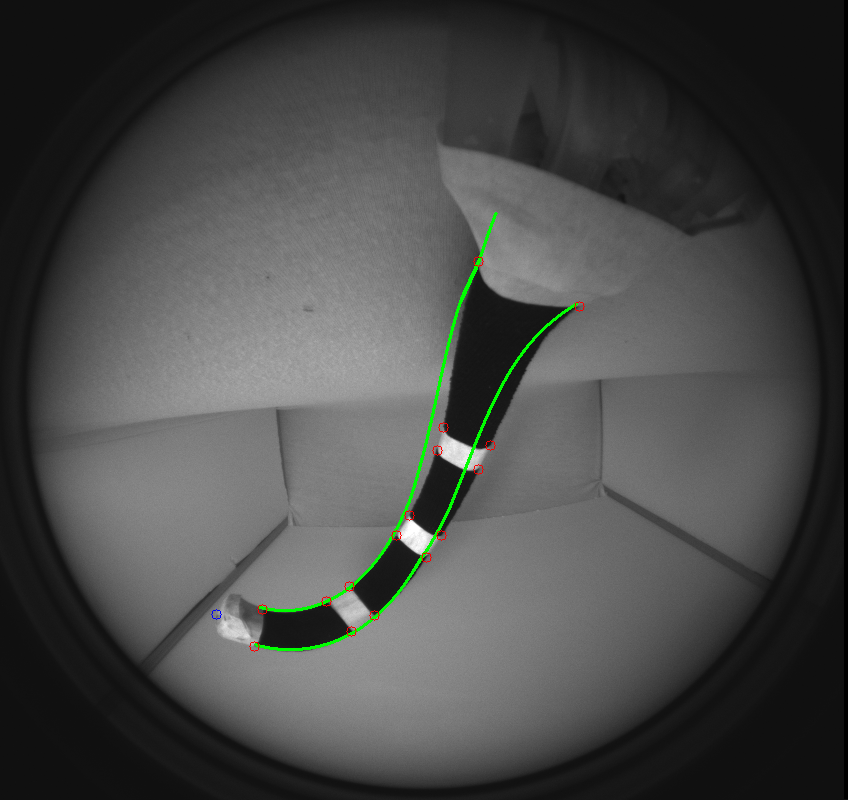}
     \end{subfigure}
      
      \hfill
      
     \begin{subfigure}[b]{0.19\textwidth}
         \centering
         \includegraphics[width=\textwidth]{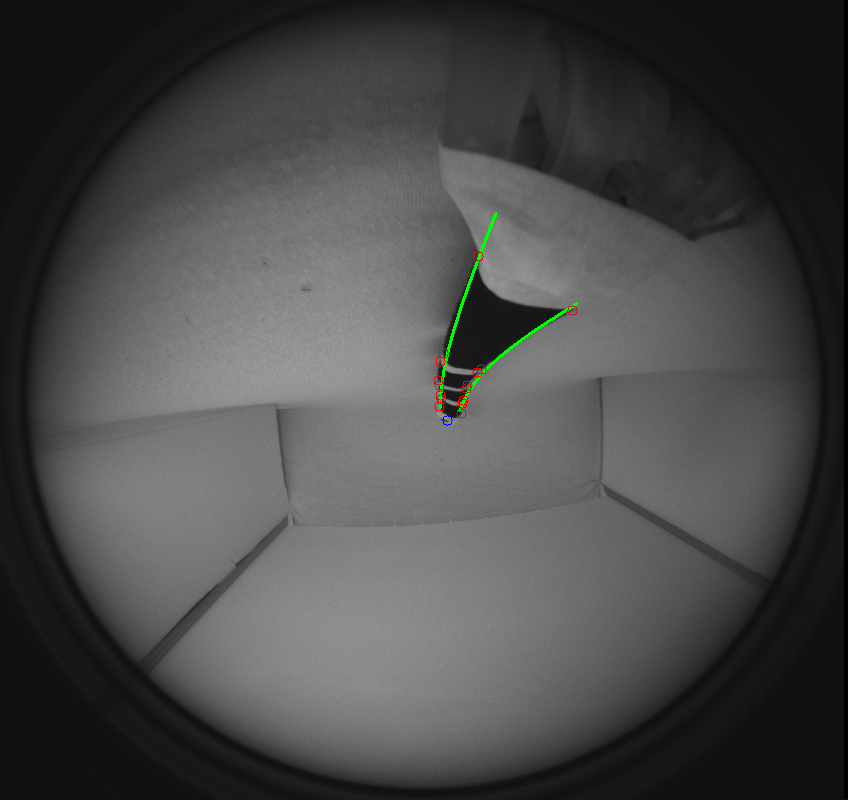}
         \caption{}
     \end{subfigure}
     \begin{subfigure}[b]{0.19\textwidth}
         \centering
         \includegraphics[width=\textwidth]{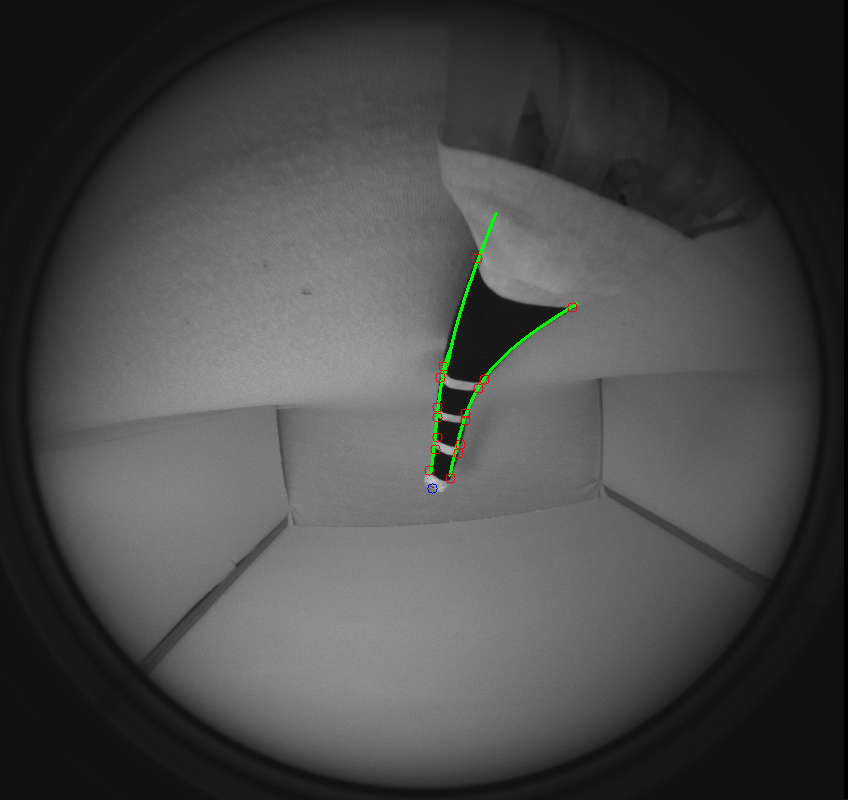}
         \caption{}
     \end{subfigure}
     \begin{subfigure}[b]{0.19\textwidth}
         \centering
         \includegraphics[width=\textwidth]{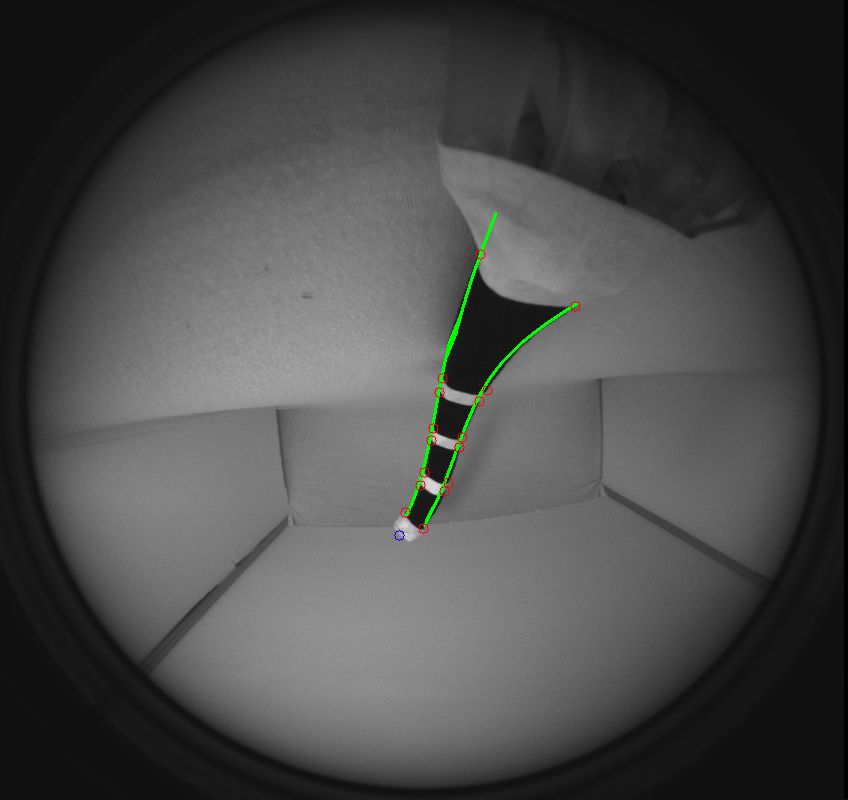}
         \caption{}
     \end{subfigure}
     \begin{subfigure}[b]{0.19\textwidth}
         \centering
         \includegraphics[width=\textwidth]{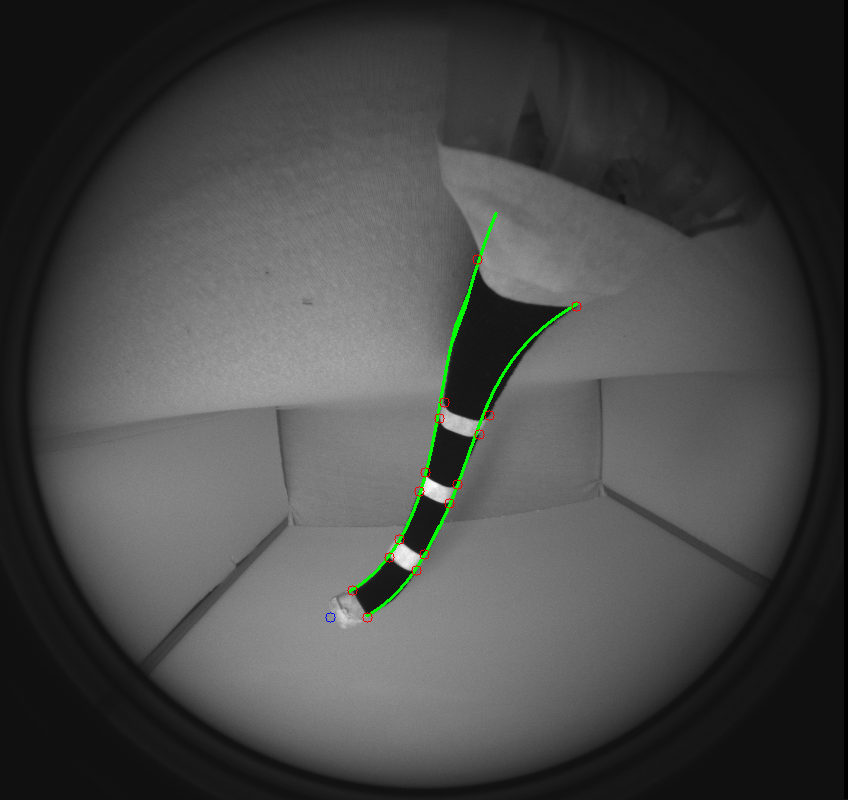}
         \caption{}
     \end{subfigure}
      \begin{subfigure}[b]{0.19\textwidth}
         \centering
         \includegraphics[width=\textwidth]{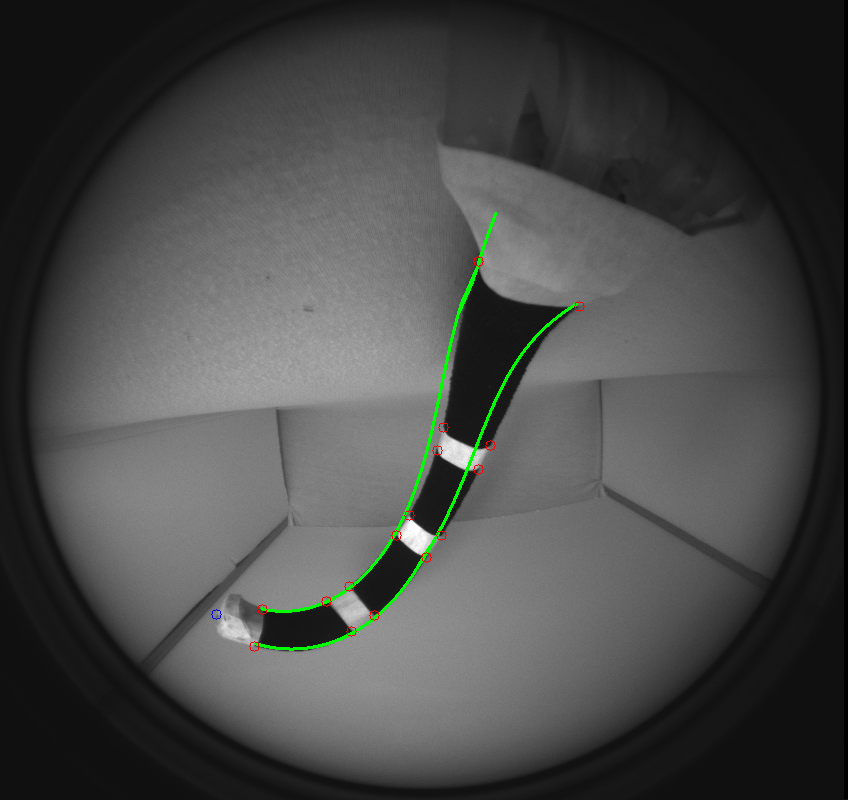}
         \caption{}
     \end{subfigure}

        \caption{Images of the soft robot with various configurations along with the projection of the estimated shape, in green, using (above) 2-segment constant strain basis and (below) quadratic basis functions.
        The red circles are the detected marker points on the image, and the blue circle is the projection of the magnetic sensor reading onto the image space.}
        \label{fig:example_3}
\end{figure*}

\section{Model of the Soft Continuum Arm Projection }\label{sec:background}
\subsection{Forward Model}
The shape of the SCA can be described by a position vector $\mathbf{p}(s)\in\mathbb{R}^3$ and a rotation frame $\mathbf{R}(s)\in SO(3)$ at each cross-section $s\in[0,L]$ along its length $L$.
For convenience, the position and orientation are expressed in a compact way by joining them into a single matrix in the special Euclidean group $SE(3)$ \cite{Lyn17} $$\mathbf{X}(s) = \begin{bmatrix} 
\mathbf{R}(s) & \mathbf{p}(s)\\
\mathbf{0} & 1
\end{bmatrix}\in SE(3).$$ 

The position and orientation of the center-points on the SCA's cross-sections evolve, with respect to its length parameter $s$, according to 
\begin{align}
    \frac{\partial\mathbf{p}(s)}{\partial s}&=\mathbf{R}(s)\mathbf{q}(s)\\
    \frac{\partial\mathbf{R}(s)}{\partial s}&=\mathbf{R}(s)[\boldsymbol{\kappa}(s)]
\end{align}
where $\boldsymbol{q} \in \mathbb{R}^3$ is a vector that contains the stretching/shearing strains, $\boldsymbol{\kappa} \in \mathbb{R}^3$ contains the bending/twisting strains, and $[\,\cdot\,]$  is the usual mapping of a vector in $\mathbb{R}^3$ to a skew-symmetric matrix in $\mathfrak{so}(3)$, the real vector space of $3$ by $3$ skew-symmetric matrices.


This can also be written compactly as
\begin{equation}
    \frac{\partial{\mathbf{X}(s)}}{\partial{s}} = \mathbf{X}(s)\mathbf{\Omega}(s),
    \label{eqn:diff_eqn}
\end{equation} 
$$
\mathbf{\Omega} = \begin{bmatrix}
[\boldsymbol{\kappa}] & \mathbf{q} \\ 0 & 0
\end{bmatrix}\in se(3).
$$
It is observed that knowing $\mathbf{\Omega}(s)$ and the initial pose at the base $\mathbf{X}_0$ is enough to reconstruct the full shape of the robot by integrating (\ref{eqn:diff_eqn}).

Given $\mathbf{\Omega}(s)$ and an initial base pose $\mathbf{X}_0$, we would like to obtain the projection of the SCA onto the camera's image sphere. This is achieved in two steps: integrating equation (\ref{eqn:diff_eqn}) to obtain the shape $\mathbf{X}(s)$ in 3D space, then projecting the 3D positions to the camera's image sphere. 

Various methods could be applied to integrate (\ref{eqn:diff_eqn}), simplest of which is to discretize the strains $\mathbf{\Omega}(s)$ into a set of $K$ piecewise constant strains $\lbrace\mathbf{\Omega}(s_k)|s_k\in\mathcal{Z_D}\rbrace$, where $\mathcal{Z_D}:=\lbrace s_1, \cdots, s_K \rbrace$.
After discretizing, it is possible to apply a simple first order integrator
\begin{equation}
\mathbf{X}(s_{k+1}) = \mathbf{X}(s_k) e^{(s_{k+1}-s_k)\Omega(s_k)}.
\end{equation}
This approach might suffer from drifting, especially if not enough discretization points are considered. More sophisticated approaches for integrating on Lie groups are the Crouch–Grossman method and  Munthe–Kaas method \cite{Par05}. Regardless of which method is used, the result of the integration is expressed as $\boldsymbol{\mathcal{X}}(s;\mathbf{X}_0)$.
\subsection{Camera Model}
We consider the case where the camera is fixed in proximity to the base of the SCA. To capture the entire workspace of the SCA, a fisheye camera is used. We note here that the shape reconstruction method presented in this paper does not depend on the type of camera or model used and can be applied to any calibrated camera that is placed such that the SCA is in its field of view. 

Without loss of generality, an image sphere is considered rather than an image plane since a wide-angle camera is considered \cite{Sca06}. The projection of a point $\mathbf{x} \in \mathbb{R}^3$ onto the image sphere centered at the origin is given by
\begin{equation}
    \mathcal{P}(\mathbf{x}) = \frac{\mathbf{x}}{\|\mathbf{x}\|}.
\end{equation}
Given a point on an image $[u,v]^T$, its corresponding image sphere projection can be obtained through the following equations
\begin{equation}
    \begin{bmatrix}
    \Bar{u}\\\Bar{v}
    \end{bmatrix}
    = \mathbf{A}
    \begin{bmatrix}
    u\\v
    \end{bmatrix} + \mathbf{c}
\end{equation}
\begin{equation}
    \mathcal{P}(\mathbf{x}) =\lambda^{-1} \begin{bmatrix}
    \Bar{u} \\ \Bar{v} \\ g(\Bar{u},\Bar{v})
    \end{bmatrix}
\end{equation}
where $g(\cdot)$ is a function that depends on the distance of an image point from the image center $\rho= \sqrt{\Bar{u}^2+\Bar{v}^2}$,
\begin{equation}
    g(\Bar{u},\Bar{v}) = a_0 + a_2\rho^2 + a_3\rho^3 + a4\rho^4,
\end{equation}
and $\lambda= \sqrt{\Bar{u}^2+\Bar{v}^2 + g(\Bar{u},\Bar{v})^2}$ is a normalizing scalar.
The coefficients $\lbrace \mathbf{A}, \mathbf{c}, a_0, a_2, a_3, a_4 \rbrace$ are dependent on the fisheye lens that is used and can be estimated through a calibration process \cite{Sca06}.

\subsection{SCA Projection onto the Camera}
In this work, knowledge of the SCA's geometry is utilized by considering the envelope of its projection onto the image (the green curves in Figure \ref{fig:example_3}).
The explicit equations for this envelope are dependent on the geometry of the SCA being used, thus explicit equations are not presented. However we generally denote for the right and left boundaries of the projected envelope as $\mathcal{P}_r(\mathbf{X}(s)), \mathcal{P}_l(\mathbf{X}(s))$, respectively.

\section{Vision-Based Shape Sensing}\label{sec:shape_recons}
This section proposes a shape sensing method that utilizes a curvature based parametrization of the SCA. First, the parametrization is introduced (a similar parametrization was introduced in \cite{Ren20}), then an optimization method that utilizes this parametrization to estimate the robot's shape is proposed.
For simplicity, an inextendable/unshearalble SCA is assumed, thus $\mathbf{q} = [1, 0, 0]^T$.
\subsection{Parameterizing $\boldsymbol{\kappa}$}\label{sec:parameterizing}
Although $[\boldsymbol{\kappa}(s)]$ lives, without further assumptions, in the space of continuous function with values in $\mathfrak{so}(3)$, for a specific SCA we assume
it lives in a finite dimensional function space and cannot have any arbitrary shape. In other words, the possible curvature profiles that the SCA adheres to can be expressed as a linear combination of a set of basis functions $\mathbf{\Phi} := \lbrace\Phi_i(s)\in \mathbb{R}^3| i = 1,\cdots,N\rbrace$
\begin{equation}
    \boldsymbol{\kappa}(s;\mathcal{A}) = \sum_{i=1}^N a_i \Phi_i(s)
\end{equation}
where $\mathcal{A}:=\lbrace a_i \in \mathbb{R}| i= 1,\cdots, N\rbrace=\mathbb{R}^N$ is the set of coefficients corresponding to the basis functions $\Phi_i(s)$. 
Applying this to represent functions valued in the Lie algebra $\mathfrak{se}(3)$ of $SE(3)$ results in
\begin{equation}
    \mathbf{\Omega}(s;\mathcal{A})=\sum_{i=1}^N a_i \Tilde{\Phi}_i(s)
\end{equation}
\begin{equation}
    \Tilde{\Phi}_i(s) = \begin{bmatrix}
[\Phi_i]_\times & \mathbf{q} \\ 0 & 0
\end{bmatrix}.
\end{equation}
Choosing specific bases, we can recover using this viewpoint some of the representations used in the literature, for example:
\begin{itemize}
    \item Constant Curvature 
$$\mathbf{\Phi}(s) := \left\lbrace 
\begin{bmatrix} 1\\0\\0 \end{bmatrix},
\begin{bmatrix} 0\\1\\0 \end{bmatrix},
\begin{bmatrix} 0\\0\\1 \end{bmatrix}
\right\rbrace,$$
    \item $N$ Piecewise Constant Curvatures
$$\mathbf{\Phi}(s) := \left\lbrace 
\begin{bmatrix} h_j\\0\\0 \end{bmatrix},
\begin{bmatrix} 0\\h_j\\0 \end{bmatrix},
\begin{bmatrix} 0\\0\\h_j \end{bmatrix} \Big | j = 1,\cdots,N
\right\rbrace,$$
$$ h_{j}(s) = \begin{cases}
               1 & s_j \leq s < s_{j+1}\\
                0 & \text{otherwise}
    \end{cases}$$
    
    \item Linear Curvature
$$\mathbf{\Phi}(s) := \left\lbrace 
\begin{bmatrix} 1\\0\\0 \end{bmatrix},
\begin{bmatrix} 0\\1\\0 \end{bmatrix},
\begin{bmatrix} 0\\0\\1 \end{bmatrix},
\begin{bmatrix} s\\0\\0 \end{bmatrix},
\begin{bmatrix} 0\\s\\0 \end{bmatrix},
\begin{bmatrix} 0\\0\\s \end{bmatrix}
\right\rbrace.$$
\end{itemize}
Other basis functions may also be used, such as higher order polynomials or trigonometric functions. With this formulation, it is possible to analyze the performance of various basis functions in capturing the space of shapes that a specific SCA can perform, or perhaps learn a basis that best describes the shapes of a SCA. However this is out of the scope of this study. 

To reflect this parametrization, the integral of equation (\ref{eqn:diff_eqn}) is written as $\boldsymbol{\mathcal{X}}(s;\mathcal{A},\mathbf{X}_0)$.
This formulation will be used in Section \ref{sec:method} to find the coefficients that best optimize for a specified cost function. 

\subsection{Optimization} \label{sec:method}
Given an
image $\mathcal{I}$ 
and a basis $\mathbf{\Phi}$ that can accurately express the possible curvature profiles of a SCA, the full shape of the SCA can be reconstructed through finding the set of coefficients $\mathcal{A}$ that minimize a suitable cost function $f(\cdot)$,
\begin{equation}
    \hat{\mathcal{A}} = \arg\min_{\mathcal{A}} f(\mathcal{A};\mathcal{I},\mathbf{X}_0).
    \label{eqn:optimization}
\end{equation}

From the image $\mathcal{I}$, 
we assume it is possible to extract image coordinates of the right and left boundary edges of the SCA's projection, $\mathbf{y}_r(s_m;\mathcal{I})$ \& $\mathbf{y}_l(s_m;\mathcal{I})$,
for a set of sample points $\mathcal{Z_I}:=\lbrace s_1, \cdots, s_M \rbrace$ on the SCA. Such task is possible using computer vision methods such as Mask RCNN \cite{KHe17}, 
applying such methods is out of the scope of this study.

To quantify the fit between the estimated shape and observed shape, we take the square root error between $\mathbf{y}_r(s_m;\mathcal{I})$ \& $\mathbf{y}_l(s_m;\mathcal{I})$ and the boundary lines $\mathcal{P}_r$  \& $\mathcal{P}_l$ of the estimated shape 
for a given set of coefficients $\mathcal{A}$,

\begin{multline}\label{eqn:cost_function}
    f(\mathcal{A};\mathcal{I},\mathbf{X}_0) = \\
    \sum_{m=1}^M \sum_{k\in\lbrace r,l \rbrace} \|\mathbf{y}_k(s_m;\mathcal{I}) - \mathcal{P}_k(\boldsymbol{\mathcal{X}}(s_m;\mathcal{A},\mathbf{X}_0))\|^2_\mathbf{w},
\end{multline}
where $\|\cdot\|^2_\mathbf{w}$ is a weighted sum of squares with $\mathbf{w}$ being a vector of (strictly positive) weights.
Equation (\ref{eqn:optimization}) can be solved using a nonlinear least squares solver such as the Trust Region Reflective Non-linear Algorithm \cite{Bra99}.

\section{Results and Discussions}\label{sec:results}
In this section, results are provided for implementing vision-based reconstruction for our specific SCA, which is the $BR^2$ manipulator \cite{Uppalapati2021}. The $BR^2$ SCA consists of three parallel combinations of fiber reinforced actuators \cite{singh2017constrained} that can bend, and twist (clockwise and counterclockwise) respectively. A spiral deformation mode can be obtained with this robot by pressurizing the bending and twisting tubes. For these experiments, the length of the arm was confined to $287$ $mm$ and the diameter to $24$ $mm$. 
Since our robot can twist along its length and only bend in one direction, we have $\boldsymbol{\kappa} = [\kappa_t, \kappa_b, 0]^T$, where $\kappa_t$ and $\kappa_b$ are associated with the twisting and bending strains, respectively. 

A fisheye camera with $160$ degree angle of view was fixed to the base of the SCA. The camera was calibrated using the MATLAB toolbox OCamCalib \cite{Sca06}. 
As in \cite{Cab14,Rei12b,Rei13}, multiple white visual markers are fixed on the SCA, as shown in Figure \ref{fig:exp_setup}. Although the proposed method can be implemented without the markers, they allow for a consistent evaluation for the method across the SCA's workspace. 
The marker positions on the image were manually identified. While this can be automated, in this work we desire to evaluate the accuracy of shape estimation without the influence of errors coming from the computer vision system.

To evaluate the accuracy of the shape estimator, an electromagnetic tracking system (Patriot SEU, Polhemus) was used to cross validate the pose measurements of the SCA's end tip, its position accuracy was found to be around $1 mm$. 
It is important to note that equation (\ref{eqn:optimization}) does not use the magnetic sensor data and only minimizes the reprojection error, the error between the projection of the estimated shape and the observed projection. 
The data from the magnetic sensor was only used to evaluate the accuracy of the tip pose estimate obtained from equation (\ref{eqn:optimization}).

\begin{figure}[t]
    \centering
    \includegraphics[width=0.48\textwidth]{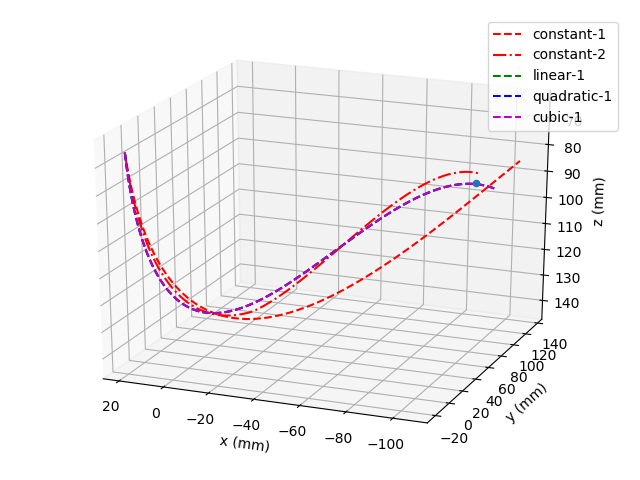}
    \caption{Estimated shape of the SCA using various basis functions. The blue point is the SCA's tip position obtained from the magnetic sensor.}
    \label{fig:3d_curve}
\end{figure}

\begin{figure}[h]
    \centering
    \begin{subfigure}[b]{0.24\textwidth}
        \includegraphics[width=\textwidth]{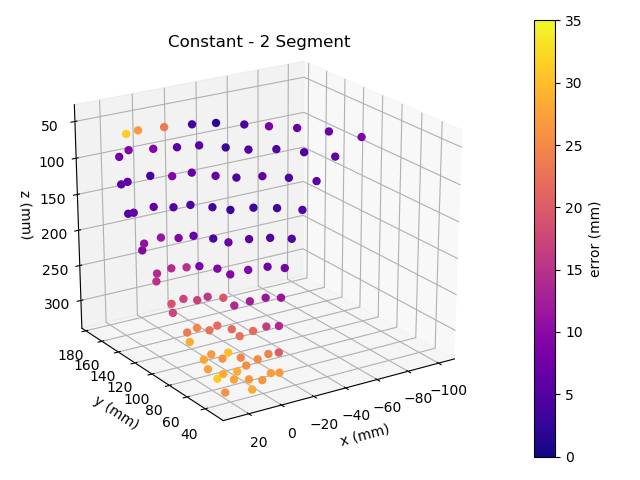}
        \caption{}
    \end{subfigure}
    \begin{subfigure}[b]{0.24\textwidth}
        \includegraphics[width=\textwidth]{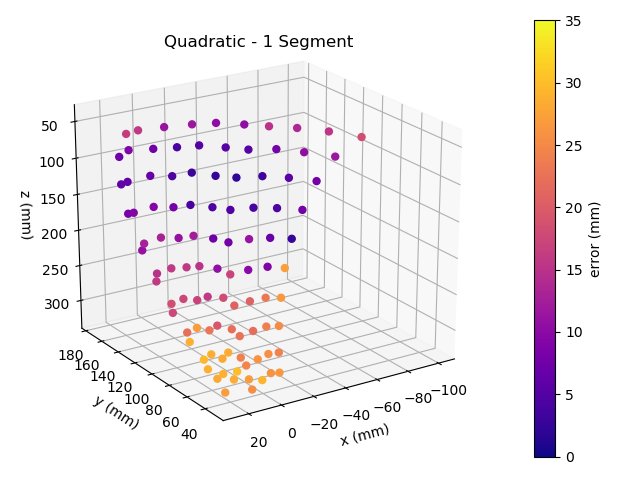}
        \caption{}
    \end{subfigure}
        \begin{subfigure}[b]{0.24\textwidth}
        \includegraphics[width=\textwidth]{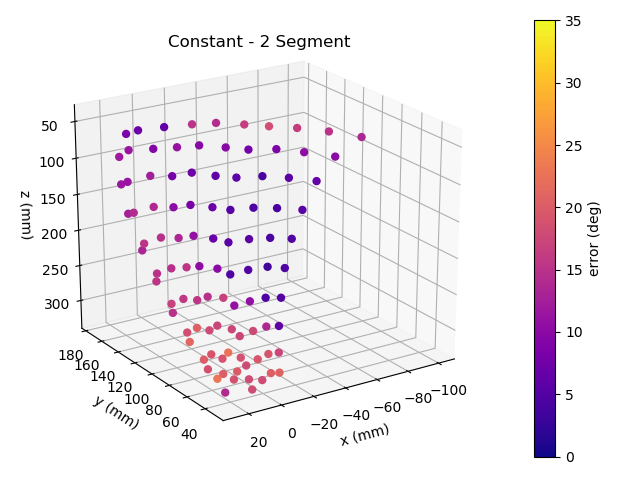}
        \caption{}
    \end{subfigure}
        \begin{subfigure}[b]{0.24\textwidth}
        \includegraphics[width=\textwidth]{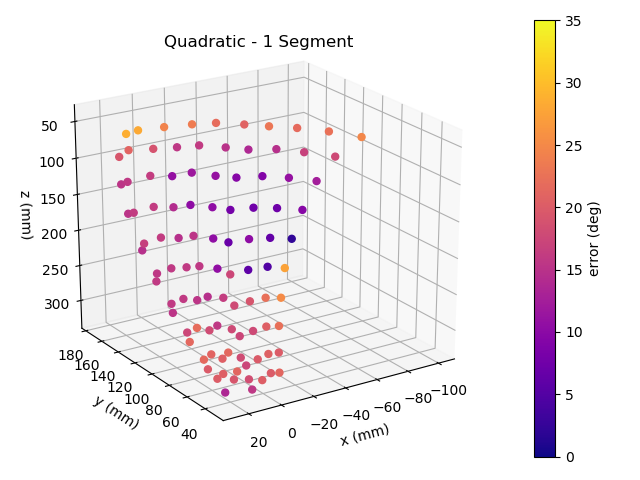}
        \caption{}
    \end{subfigure}
    \caption{3D position samples from the SCA's workspace with the corresponding error in (a,b) the tip position and (c,d) tip direction using a 2-segment constant strains basis and a quadratic basis.}
    \label{fig:spatial_error}
\end{figure}

\subsection{Experimental Procedure}
A series of experiments were conducted to evaluate the accuracy of the proposed shape sensing method.
In these experiments, various configurations were tested by applying different pressures for the bending and the twisting actuators. The bending and twisting actuator were given $10$ pressures each ranging from $0$ to $25$ psi and $0$ to $20$ psi, respectively. From the combinations of these pressures, 100 sample configurations were obtained for half of the robot's workspace. At each sample, an image was captured and readings from the magnetic sensor were collected.
These measurements were preprocessed to find unknown parameters such as the pose of the SCA's base and the relative transformation between the camera frame and the magnetic source frame.

\subsection{Pose of SCA's Base}
The SCA's base with respect to the camera's frame, $\mathbf{X}_0$, is unknown. 
Here we present an approach that utilizes the same optimization framework and provides an accurate estimate of the transformation.
The relative transformation between the base of the SCA and the camera was obtained by finding the pose that minimizes the error between the observed projection of the base, $\mathbf{y}_r(0;\mathcal{I})$ \& $\mathbf{y}_l(0;\mathcal{I})$, and the projection of the estimated base throughout the whole set of images $\lbrace\mathcal{I}_i|i=1,\dots,N\rbrace$
\begin{equation}
    \hat{\mathbf{X}}_0 = \arg\min_{\mathbf{X}_0} \sum_{i} \sum_{k\in\lbrace r,l \rbrace}
    \|\mathbf{y}_{k}(0;\mathcal{I}_i) - \mathcal{P}_k(\mathbf{X}_0)\|^2.
    \label{eqn:optimization_init}
\end{equation} 
Once the initial pose $\hat{\mathbf{X}}_0$ was found, it was fixed and used to solve (\ref{eqn:optimization}).

\subsection{Pose Data from the Magnetic Sensor}
The pose obtained from the magnetic sensor, $\mathbf{X}_{sen}^{mag} \in SE(3)$, is with reference to a magnetic source inside the lab. However the estimate obtained from equation (\ref{eqn:optimization}) is with respect to the camera frame. To be able to cross validate the estimated poses with the magnetic sensor readings, we need to find the relative transformation between the camera frame and the magnetic source frame, $\mathbf{X}_{mag}^{cam}$, in order to represent the data with respect to the camera frame.
Furthermore, the magnetic sensor does not align exactly with the SCA's tip, therefore we need to obtain the relative transformation between the magnetic sensor and the SCA's tip, $\mathbf{X}_{tip}^{sen}$.

An estimate of these relative transformations were obtained by minimizing the error between the observed projection of the tip, $\mathbf{y}_r(L;\mathcal{I})$ \& $\mathbf{y}_l(L;\mathcal{I})$, and the projection of the transformed sensor readings throughout the whole set of images
\begin{multline}
    \hat{\mathbf{X}}_{mag}^{cam}, \hat{\mathbf{X}}_{tip}^{sen} =  \arg\min_{\mathbf{X}_m^c,\mathbf{X}_t^s} \\\sum_{i} \sum_{k\in\lbrace r,l \rbrace} \|\mathbf{y}_k(L;\mathcal{I}_i) - \mathcal{P}_k(\mathbf{X}_m^c\mathbf{X}^{mag}_{sen_i}\mathbf{X}^s_t))\|^2.
    \label{eqn:optimization_mag}
\end{multline}
The pose $\mathbf{X}^s_t$ was constrained to a subset of $SE(3)$, translations along the x and z-axis and rotation around the x-axis. We numerically verified that equation (\ref{eqn:optimization_mag}) always converges locally to the same values.
Once the transformations were estimated, the magnetic sensor readings were transformed to the corresponding tip position with respect to the camera frame. This was then used to measure the error in the estimated tip position and direction.


\begin{table*}[!h]
\begin{center}
\caption{Comparison in the soft robot tip position and direction estimation errors for various strain basis functions}
\label{tab:2}
\setlength\tabcolsep{0pt} 

\begin{tabular*}{\textwidth}{@{\extracolsep{\fill}} ccc cccccccc}
\toprule
     \multirow{3}{*}{Order}  & \multirow{2}{*}{Segments} &  \multicolumn{4}{c}{Error in entire workspase} &  \multicolumn{4}{c}{Error in region A}\\ 
\cmidrule{3-6}
\cmidrule{7-10}
     & (Parameters) &
     $\mu(E_1)\pm\sigma(E_1)$ & $\max(E_1)$ & $\mu(E_2)\pm\sigma(E_2)$\tnote{c} & $\max(E_2)$ & 
     $\mu(E_1)\pm\sigma(E_1)$ & $\max(E_1)$ & $\mu(E_2)\pm\sigma(E_2)$\tnote{c} & $\max(E_2)$ \\
     & &$(mm)$ & $(mm)$ & (degree) & (degree) & $(mm)$ & $(mm)$ & (degree) & (degree) \\
\midrule
     0  & 1 (2)     & 29.3 $\pm$ 1.2 & 31.9  & 29.3 $\pm$ 6.6 & 39.8 & 29.8 $\pm$  0.9 & 31.9  & 34.9 $\pm$ 3.0 & 39.8\\
    \addlinespace
     0  & 2 (4)     & \textbf{13.0 $\pm$ 9.0} & \textbf{28.2}  & \textbf{12.2 $\pm$ 4.7} &  \textbf{21.3} & \textbf{6.1 $\pm$ 5.3} &  \textbf{28.2} & \textbf{9.9 $\pm$ 3.1} & \textbf{16.5}\\
    \addlinespace
     1  & 1 (4)     & 14.9 $\pm$ 9.2 & 30.4  & 15.7 $\pm$ 5.3 & 33.2 & 7.7 $\pm$ 7.2 & 30.4   & 13.5 $\pm$ 6.4 & 33.2\\
     \addlinespace
     2  & 1 (6)     & 15.1 $\pm$ 9.1 & 30.4  & 15.6 $\pm$ 5.3 & 33.2 & 8.2 $\pm$ 7.7 & 30.4   & 13.6 $\pm$ 6.4 & 33.2\\
    \addlinespace
     3  & 1 (8)     & 15.1 $\pm$ 9.1 & 30.4  & 15.6 $\pm$ 5.3 & 33.2 & 8.2 $\pm$ 7.7 & 30.4  & 13.6 $\pm$ 6.4 & 33.2\\
\bottomrule
\end{tabular*}

\smallskip
\scriptsize
\begin{tablenotes}
\RaggedRight
\item Soft robot has a length $287$ $mm$ and diameter $24$ $mm$.
\item Region A is the upper half region of the workspace shown in Figure \ref{fig:spatial_error}.
\item E1: Error in the tip position.
\item E2: Error in the tip direction angle.
\end{tablenotes}
\end{center}
\end{table*}

\subsection{Results and Analysis}
The proposed method has been tested with five different basis functions for the curvature: constant, piecewise constant, linear, quadratic, and cubic functions.
The weights, in equation (\ref{eqn:cost_function}), have been chosen to increase linearly with the length of the arm (i.e. more weight is given to the end tip).
The mean, standard deviation, and maximum of the errors in the estimated tip position and direction are summarized in Table \ref{tab:2}.
For the specific SCA being used in the experiment, the basis with minimum error is a constant piecewise function with two segments with an error of $13.0$ $mm$ and $12.2$ degrees in the position and angle, respectively.
Relative to the diameter of the SCA, the estimated tip position is, on average, within approximately half of the diameter.
It is worthy to note that, for the BR$^2$, increasing the order of the polynomial to the third order or more does not improve the results, as seen in the last line of Table \ref{tab:2}.

From the spatial distribution of the errors (see Figure \ref{fig:spatial_error}) it can be observed that the region with maximum error is when the SCA is close to being fully extended. This could be due to the difficulty of getting accurate marker coordinates when the SCA is in this configuration. This region is also more sensitive to errors in the image coordinates of the detected markers, since small changes in the marker coordinates contribute to large displacements in the tip position. Another way to look at this is to observe that there is a certain configuration (outside of the robots workspace) where all the markers will get projected to the same point on the image, and thus shape reconstruction becomes difficult when this configuration is approached. 

For some applications, the upper half of the work space might be more important (i.e. when the SCA is bending more). When considering only this region, an improvement in the estimation accuracy is observed. A two-segment piecewise basis achieves an accuracy of $6.1$ $mm$ and $9.9$ degrees.

One source of error includes inaccuracies in camera calibration, especially since the fisheye lenses have high distortions on the edges compared to narrow angle lenses. This could be seen when the SCA's tip is close to the edges of the cameras view, as in Figure \ref{fig:example_3} (e). These regions have higher errors due to the lens distortions. Fine tuning the camera calibration parameters could resolve this issue. Another source of error could be in the detection of the marker positions in the image. Since the number of markers is relatively low, small errors in their image coordinates will lead to errors in the overall shape estimation. This can be resolved by considering the entire projected curve rather than only on the markers, thus improving the estimation accuracy.

\section{Conclusion}\label{sec:conclusion}
Accurate 3D shape reconstruction of a soft continuum arm is essential for applications that require accurate interactions with the environment. The lack of accurate cost effective solutions for this problem makes it difficult to deploy autonomous SCAs in real world scenarios \cite{Uppalapati2020a}. In this paper, a vision based approach for estimating the SCA's shape was proposed. The method utilized a fish-eye camera attached to the base of the SCA that is able to see its whole workspace. A generic curvature based representation of the SCA's shape was used to efficiently optimize for the shape that reduces the reprojection error. 
Results show the effectiveness of the proposed method, which gave results of accuracy less then the SCA's diameter. This margin of error is acceptable for most applications. 

Out of the tested basis functions, the best performance for the BR$^2$ was achieved with two constant strain segments. The performance of other types of basis functions can be analyzed in the future for a more comprehensive comparison
. Also, an interesting direction would be to learn a basis that describes the SCA's shape the best. 

The SCA used in this work bends only in one direction, therefore one camera was enough to capture its workspace. However for applications where a robot that bends in both direction is needed, this approach can be extended by placing multiple cameras around the base of the SCA.
One drawback of using a camera to estimate the robot's shape is its vulnerability when the SCA is occluded or partially occluded. This drawback can be dealt with by extending the proposed method to accept and fuse other sensor measurements that can be helpful, such as the internal pressures or a curvature sensor.

The work presented in this paper will be built upon to develop autonomous capabilities for SCA's that would be useful in applications such as fruit harvesting \cite{Uppalapati2020a}, robotic care-giving, and surgery. 
More specifically, it can be applied as a feedback loop for controlling the SCA's end-effector to a desired position. Also, it can be used for applications where the whole shape of the SCA is needed, such as detecting obstacles or estimating the contact forces applied on the SCA.
To deploy the proposed system in real world application, realtime performance is needed. 
It is possible to optimize the proposed method for high-speed computation using realtime programming language and by utilizing parallelization.

\bibliographystyle{IEEEtran}
\bibliography{references,GKref}

\end{document}